\documentclass{ifacconf}

\usepackage{graphicx}      
\usepackage{natbib}        
\usepackage{algorithm}
\usepackage{algpseudocode}
\usepackage{amsmath}
\usepackage{amssymb}
\usepackage{subcaption}
\begin{document}
\begin{frontmatter}

\title{Automating Box Folding: Sequence Extraction and Ranking Methodologies} 

\author[First]{Giuseppe Fabio Preziosa} 
\author[First]{Davide Ferloni} 
\author[First]{Andrea Maria Zanchettin}
\author[First]{Marco Faroni}
\author[First]{Paolo Rocco}

\address[First]{The Authors are with Politecnico di Milano, Dipartimento di Elettronica,Informazione e Bioingegneria (DEIB), Piazza Leonardo da Vinci 32, 20133, Milano (Italy). (e-mail: giuseppefabio.preziosa, davide.ferloni, andreamaria.zanchettin, marco.faroni, paolo.rocco)@polimi.it}


\begin{abstract}                
Box folding represents a crucial challenge for automated packaging systems. This work bridges the gap between existing methods for folding sequence extraction and approaches focused on the adaptability of automated systems to specific box types. An innovative method is proposed to identify and rank folding sequences, enabling the transformation of a box from an initial state to a desired final configuration. The system evaluates and ranks these sequences based on their feasibility and compatibility with available hardware, providing recommendations for real-world implementations. Finally, an illustrative use case is presented, where a robot performs the folding of a box.
\end{abstract}

\vspace{0.8cm}

\end{frontmatter}

\section{Introduction}
\vspace{-0.2cm}
In the context of modern logistics and the growing demand for efficient goods handling, automated box folding has become an essential process for many industries. Beyond logistics, any company that produces goods (across a wide range of sectors) may need to automate its packaging operations to keep up with increasing production volumes. As production scales up, the need to automate the packaging process—from the object-inserting phase to the folding of boxes—becomes ever more critical, not only to reduce labor costs but also to increase throughput and maintain consistency in product handling. While recent research primarily focuses on aspects such as filling and arranging objects to optimize space with \cite{packing_objects1}, improving package stability from \cite{stable_packing}, and arranging irregular objects with \cite{irregular_objects}, relatively few works address the actual box folding process.  

The problem of determining how to fold a carton is typically approached from two opposite perspectives. One approach focuses on the theoretical aspect, deriving all possible folding sequences without considering the type of machine possibly executing them, leaving the selection of an implementable sequence to human intuition or expertise. The other approach develops solutions limited to specific types of cartons and tailored to the available hardware, sacrificing flexibility and adaptability. In contrast, this work aims to address the problem comprehensively, bridging the gap between these two extremes. By generalizing the process for any carton design and introducing metrics to evaluate folding sequences, this research provides a systematic framework that not only identifies feasible sequences but also supports their implementation in diverse industrial settings.

\begin{figure}
\begin{center}
\includegraphics[width=8.3cm]{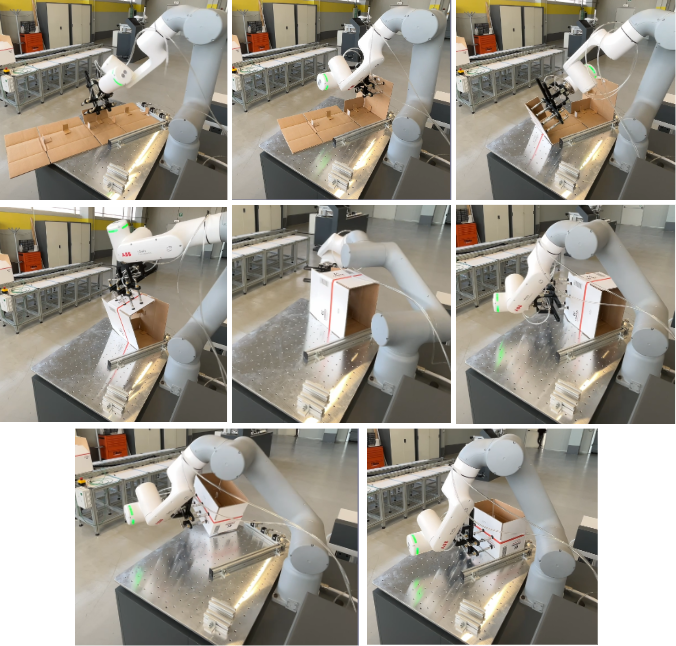}    
\vspace{-0.3cm}
\caption{Folding steps of the implemented sequence extracted by the proposed algorithm.} 
\label{fig:folding}
\end{center}
\end{figure}

\section{Related Works}

The problem of automated carton folding has attracted significant attention in past research, though interest in this area seems to have waned in recent years. However, with the growing demand for flexible and efficient packaging systems, it is still important to contribute to this field by exploring innovative solutions and methodologies.
Several works have investigated carton folding, focusing on two main research domains. The first involves \emph{how to represent and model} a box structure, while the second one focuses on \emph{robotic manipulation strategies}, ranging from fixture-based approaches to the use of advanced automation tools.

\subsection{Box Representation and Kinematic Modeling} \label{Box Representation and Kinematic Modeling}
Initial works, such as those by \cite{mobilityOfFoldable}, \cite{configurationTransformation}, and \cite{kinematicAndMobility}, conceptualize cartons as \emph{metamorphic mechanisms}. Panels are represented as links, while creases act as revolute joints. This framework captures the evolving mobility of cartons during the folding process. To understand and model the connections between joints (creases) and links (panels), \cite{configurationTransformation} introduces the concepts of adjacency and hereditary connectivity matrices. These matrices describe the relationships among components and are further reorganized into directed graphs, as shown in \cite{configurationTransformation} and \cite{kinematicAndMobility}, using graph theory to analyze kinematic paths.

Modeling also requires specifying the dimensions of panels and creases to ensure accurate representation. \cite{configurationTransformation} and \cite{kinematicAndMobility} employ parameters such as panel depth, thickness, breadth, and crease length and calculate the geometric center of each panel. These centers serve as fixed reference points for applying rotations and translations, ensuring precise folding movements.

To describe the carton mobility during folding, \cite{configurationTransformation}, \cite{kinematicAndMobility}, and \cite{mobilityOfFoldable} utilize the Grübler-Kutzbach mobility criterion by \cite{mobilityCriterion} to calculate the degrees of freedom available at different folding stages. Additionally, \cite{mobilityOfFoldable} and \cite{kinematicAndMobility} employ Screw Theory by \cite{motionRigidBody} to further analyze motion, offering a detailed understanding of kinematic constraints and mobility transitions.

Kinematic modeling also defines the constraints that govern allowable movements and folding angles to maintain structural consistency. \cite{configurationTransformation}, \cite{kinematicAndMobility}, and \cite{reconfigurableRoboticSystem} explore these constraints, addressing the range of permissible movements and the specific folding angles required. They also examine folding sequences, emphasizing that certain folds must occur in a defined order to prevent overlaps or collisions.
Path planning complements kinematic modeling by enabling collision-free folding sequences. \cite{motionPlanningApproach} introduces a path planning approach that uses the configuration space (C-space from \cite{CSpace}) to compute feasible paths. This ensures each fold aligns with spatial constraints while maintaining the carton structural integrity.

\subsection{Robotics for Box Handling} \label{Robotics for Box Handling}
The key requirements for robotic systems in carton handling include adaptability to different carton shapes, reliability in repeated operations, and speed to meet industrial throughput demands. In literature, two primary approaches have emerged to address the diverse requirements of packaging automation. The first relies on fixture-based systems, which simplify carton handling by using specialized fixtures to stabilize and position the carton, reducing the need for complex robotic movements. For example, \cite{motionPlanningApproach} employs a motion planning approach that uses fixtures to manage the precise positioning of cartons, enabling the robotic arm to perform basic movements with reliable results. Similarly, \cite{lowCostAutomation} presents a low-cost automation solution for folding grape packaging cartons, designed for small-batch production and offering a cost-effective solution. While both works present fixture-based systems as highly effective for standardized carton folding, they face significant challenges in handling variability in carton shapes and sizes, as each new carton design requires a complete fixture redesign, lacking flexibility for different carton styles.

The second approach focuses on mechanically complex robotic systems that use mechanical fingers and articulated joints to handle intricate carton folding. Unlike fixture-based setups, these systems perform detailed folding operations, allowing for adaptability to diverse carton styles and complex designs. \cite{reconfigurableRoboticSystem} introduces a reconfigurable robotic system designed with modular automatic fingers that can manipulate each carton independently, enabling adaptability to any carton type. Similarly, \cite{confectioneryIndustry} presents an automated solution tailored specifically for the confectionery industry, featuring articulated joints and modular components. Despite their versatility, mechanically complex systems involve higher costs and require significant maintenance efforts. The challenging upkeep and reconfiguration processes can result in extended downtime whenever a new carton design is introduced, emphasizing the trade-offs between flexibility and operational efficiency in these approaches.

\subsection{Contributions}
This study aims to bridge the gap between general folding sequence extraction and hardware-specific implementation. At its core, our approach introduces an algorithm whose aim is to identify feasible folding sequences that prioritize collision avoidance. These sequences are generated in a hardware-independent manner, focusing solely on determining viable folding paths.
Unlike previous approaches, which either develop methods tailored to specific hardware or explore folding sequence analysis from a purely theoretical standpoint, this work goes further by proposing innovative indices to evaluate and recommend the most suitable folding sequences based on the automation hardware employed.

\section{Proposed Approach}
This work tackles the challenge of automating the folding process for flat cartons by identifying and suggesting feasible folding sequences for real-world implementation. The methodology consists of three phases: representing the carton structure through kinematic modeling and connectivity analysis (Section \ref{sec:box_modeling}), systematically identifying feasible folding sequences using decision trees (Section \ref{sec:sequence_identification}), and introducing evaluation metrics to rank and prioritize sequences based on hardware compatibility (Section \ref{sec:metric_evaluation}).

\begin{figure*}[h]
\begin{center}
\begin{tabular}{cc}
{\hspace{-0.4cm}\includegraphics[width=0.5\textwidth]{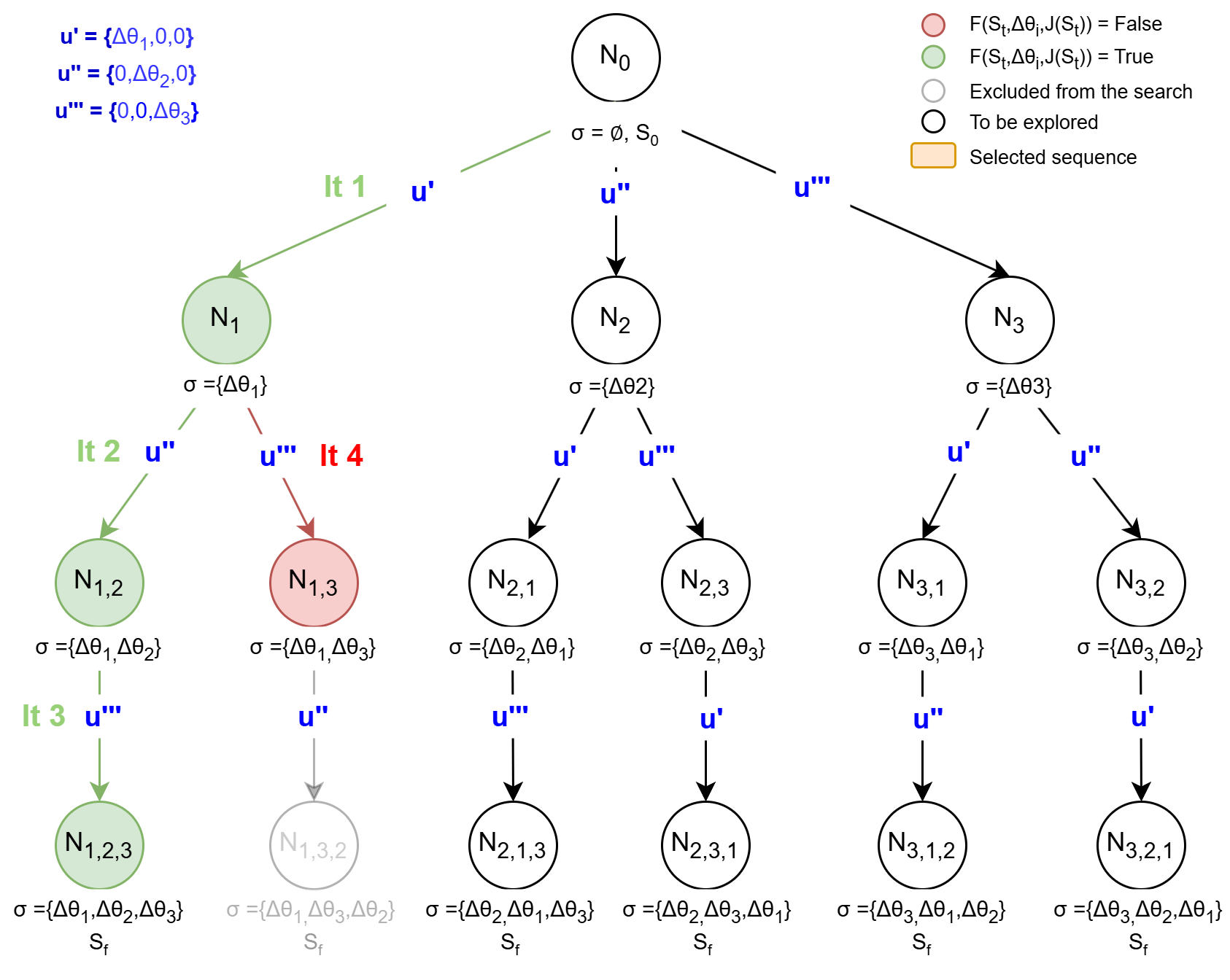}} & 
{\hspace{-0.1cm}\includegraphics[width=0.5\textwidth]{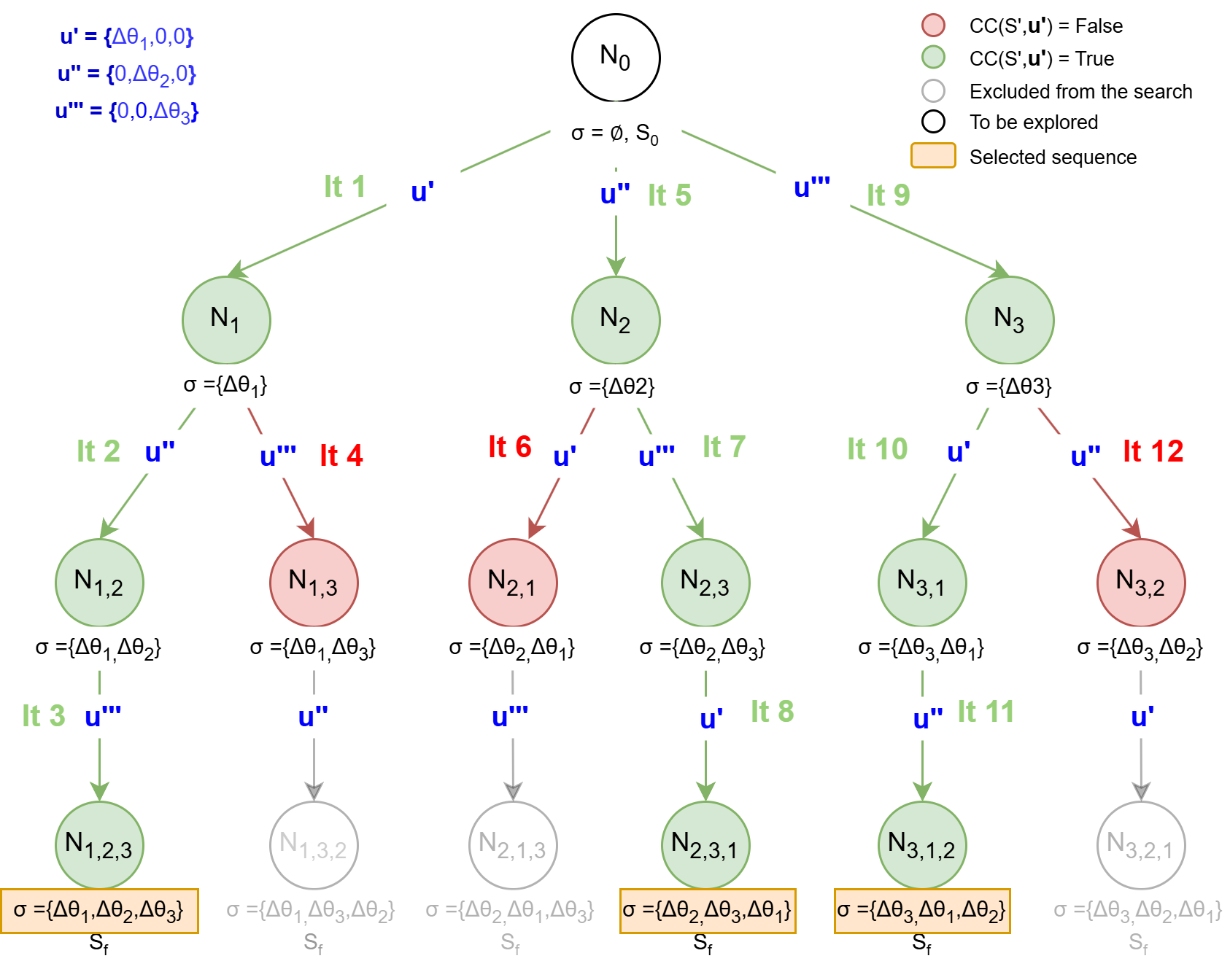}}  \\ 
\end{tabular}
\vspace{-0.4cm}
\caption{Simplified structure of a three-panel decision tree. On the left, the first five iterations of the backtracking process are shown, highlighting excluded paths and intermediate states. On the right, the completed search is depicted, with the valid sequences reaching the final states \(S_f\) highlighted in orange.}
\label{fig:tree_exp}
\end{center}
\end{figure*}

\subsection{Box Modeling and Connectivity Matrix Generation} \label{sec:box_modeling}

This phase focuses on representing the carton structure and its folding capabilities using tools from kinematic modeling discussed in Section \ref{Box Representation and Kinematic Modeling}. Following established approaches, the carton can be modeled as a mechanism composed of joints and links, where folds are treated as rotational joints and panels as rigid links. Given a box with $n$ joints, we define $\theta$ as the joint angle vector:
\begin{equation}
\boldsymbol{\theta} = [\theta_1, \theta_2, \dots, \theta_n]^T,
\end{equation}
where \( \theta_i \) denotes the rotation angle of joint \( i \).
To capture the structural relationships between panels, we introduce the connectivity matrix \( C \), defined as:
\begin{equation}
C_{ij} = \begin{cases} 
1 & \text{if panel } i \text{ influences panel } j, \\
0 & \text{otherwise.}
\end{cases}
\end{equation}
This matrix encodes both adjacency and hierarchical dependencies, allowing the identification of parent-child relationships, where the motion of a parent panel affects its child panels. The carton state can thus be compactly represented by the variable \( S \), defined as:
\begin{equation}
S = \{ \boldsymbol{\theta}, C \}.
\end{equation}
Based on this representation, the state of the carton can be further generalized using the connected graph \( G_t = (V, E_t) \), where \( V \) represents the set of panels and \( E_t \) the edges describing the connections between panels at time \( t \). The connected graph is a dynamic representation that captures changes in degrees of freedom during the folding process. Consequently, the joint angle vector \( \boldsymbol{\theta} \) and the connectivity matrix \( C \) can be expressed as functions of the connected graph \( G_t \):
\begin{equation}
\boldsymbol{\theta}, C = g(G_t)
\end{equation}
To complement the topological representation, the physical properties of each panel are incorporated. Panels are characterized by their dimensions and positions, stored in the following matrices:
\begin{equation}
D = \begin{bmatrix} h_1 & w_1 & t_1 \\
                   h_2 & w_2 & t_2 \\
                   \vdots & \vdots & \vdots \\
                   h_n & w_n & t_n \end{bmatrix}, \quad
P = \begin{bmatrix} x_1 & y_1 & z_1 \\
                   x_2 & y_2 & z_2 \\
                   \vdots & \vdots & \vdots \\
                   x_n & y_n & z_n \end{bmatrix}
\end{equation}
where \( h_i, w_i, t_i \) are the height, width, and thickness of panel \( i \), and \( (x_i, y_i, z_i) \) is its geometric center.

\subsection{Folding Sequence Identification} \label{sec:sequence_identification}

Building upon the representation formalized in Section \ref{sec:box_modeling}, the folding process is defined as a sequence of discrete actions that transition the carton from an initial state \( S_0 \) to a final state \( S_f \). The initial state \( S_0 \) is characterized by joint angles \( \theta_i^{\text{init}} \), while the final state \( S_f \) corresponds to joint angles \( \theta_i^{\text{final}} \).
At each step \( t \), for a box with \( n \) panels, an action vector \( \boldsymbol{u_t} = \{u_1, u_2, \dots, u_n\} \) is defined, where \( u_i \) represents the action applied to joint \( i \). Specifically, \( u_i \) is either a complete rotation that moves the joint \( i \) from \( \theta_i^{\text{init}} \) to \( \theta_i^{\text{final}} \) or \( 0 \). The action space \( U(S_t) \) is defined as:
\begin{equation}
\begin{aligned}
U(S_t) = 
&\Bigg\{ \boldsymbol{u_t} = \{u_1, u_2, \dots, u_n\} \mid \\
& \quad u_i = 0 \, \text{if } \theta_i^t = \theta_i^{\text{final}}, \\
& \quad \text{else } u_i \in \{0, \theta_i^{\text{final}} - \theta_i^{\text{init}}\}, \\
& \quad \sum_{i=1}^n \mathbb{I}(u_i \neq 0) = 1 \Bigg\},
\end{aligned}
\end{equation}
where \( \mathbb{I} \) is the indicator function. The formulation ensures that, at each step \( t \), only one joint is rotated from the initial to the final configuration, while all other joints remain stationary. For example, in a carton with 3 panels where all the panels can be rotated, 
\( \boldsymbol{u_t} \) at a given iteration might be 
\(\boldsymbol{u_t} = [\Delta \theta_1, 0, 0]\), 
\(\boldsymbol{u_t} = [0, \Delta \theta_2, 0]\), or 
\(\boldsymbol{u_t} = [0, 0, \Delta \theta_3]\), 
indicating that only one panel is folded at a time. The state transition is governed by the function \( f \), which updates the state \( S_t \) based on the applied action vector \( \boldsymbol{u_t} \). The updated state is given by:
\begin{equation}
S_{t+1} = f(S_t, \boldsymbol{u_t}).
\end{equation}
To validate the action \( \boldsymbol{u_t} \) from a state \( S_t \), a collision-checking function \( CC(S_t, \boldsymbol{u_t}) \) is defined. This function verifies whether the incremental rotation specified by \( \boldsymbol{u_t} \) causes any collisions. It is defined as:
\begin{equation}
CC(S_t, \boldsymbol{u_t}) = 
\begin{cases} 
\text{True}, & \text{if no collision occurs}, \\
\text{False}, & \text{otherwise.}
\end{cases}
\end{equation}
We can define a valid sequence \( \sigma_j \) as the action sequence \( \{\boldsymbol{u_t}\}_{t=0}^{k} \) that satisfies the following constraints:
\begin{equation}
\begin{aligned}
& S_{t+1} = f(S_t, \boldsymbol{u_t}), \\
& \boldsymbol{u_t} \in U(S_t), \\
& S_{t=0} = S_0, \quad S_{t=k} = S_f, \\
& CC(S_t, \boldsymbol{u_t}) = \text{True}, \quad \forall t \in \{0, \dots, k\}
\end{aligned}
\end{equation}
This formulation ensures that the carton transitions from the initial state \( S_0 \) to the final state \( S_f \) while respecting geometric and kinematic constraints. By requiring \( \boldsymbol{u_t} \) to belong to the action space \( U(S_t) \) and verifying each action through \( CC(S_t, \boldsymbol{u_t}) \), the process avoids collisions and guarantees that only valid folding actions are applied.

The problem thus becomes identifying all \( q \) feasible sequences \( \sigma_1, \dots, \sigma_q \) that satisfy Equation (9). To achieve this, we adopt a decision tree representation, where each node is defined as \( N = (S'', S', \boldsymbol{u'}) \), with \( S'' \) representing the current state of the carton, \( S' \) the parent state, such that \( S'' = f(S', \boldsymbol{u'}) \). The root node is initialized based on the initial state \( S_0 \), while successful transitions between nodes are governed by \( CC(S', \boldsymbol{u'}) = \text{True} \). This ensures that only valid action vectors are considered, avoiding collisions and respecting kinematic constraints. To explore the decision tree and identify all valid sequences, a backtracking algorithm from \cite{backtrackingAlgorithms} is employed. This algorithm systematically traverses the tree using a depth-first search strategy, exploring each path as far as possible. When a transition fails to satisfy \( CC(S_t, \boldsymbol{u}_t) = \text{True} \), the algorithm backtracks to the parent node and evaluates alternative actions. A sequence \( \sigma_j\)  is then recorded only if it leads to the final state \( S_f \). This approach guarantees that all feasible configurations are explored while efficiently pruning invalid paths. A visual representation of the tree and the search algorithm is shown in Fig. \ref{fig:tree_exp}.

\subsection{Evaluation Metrics for Ranking Folding Sequences}
\label{sec:metric_evaluation}

The innovative contribution of this work lies in the introduction of evaluation metrics designed to assess which folding sequence can be effectively implemented on specific hardware. These metrics allow for the ranking of sequences identified in the preceding phase, ensuring better alignment with the practical requirements. For a feasible sequence \( \sigma_j \) of length \( k \), expressed as \( \{\boldsymbol{u_t}\}_{t=0}^{k} \), which represents a progression of states \( S_0 \to S_1 \to \dots \to S_f \), each cost function evaluates specific aspects of the folding process associated with \( \sigma_j \). The evaluation details are presented below.

\subsubsection{Bounding Box Volume Criterion}

This criterion minimizes the cumulative spatial volume occupied by the carton during the folding process, promoting compactness and stability. For a given carton configuration $S_t$, the volume $V(S_t)$ of the smallest bounding box enclosing the carton is expressed as:
\begin{equation}
V(S_t) = l(S_t) \cdot w(S_t) \cdot h(S_t),
\end{equation}
where $l(S_t)$, $w(S_t)$, and $h(S_t)$ represent the length, width, and height of the bounding box at configuration $S_t$ respectively.
The total cost $C_{\text{vol}}(\sigma)$ for a folding sequence $\sigma$ is calculated as the cumulative volume across all intermediate configurations:
\begin{equation}
C_{\text{vol}}(\sigma) = \sum_{t=0}^{k-1} V(S_t).
\end{equation}
By minimizing $C_{\text{vol}}(\sigma)$, this criterion ensures compact folding, reduces collision risks, and ease of manipulation.

\subsubsection{Bounding Box Maximum Dimension Criterion}

This criterion minimizes the largest single dimension of the bounding box across all folding steps to improve accessibility and ensure compatibility with robotic hardware constraints, such as workspace limitations. For a given configuration $S_t$, the maximum dimension $\text{MaxDim}(S_t)$ of the bounding box is defined as:
\begin{equation}
\text{MaxDim}(S_t) = \max\big(l(S_t), w(S_t), h(S_t)\big).
\end{equation}
The total cost $C_{\text{dim}}(\sigma)$ for a sequence $\sigma$ is given by:
\begin{equation}
C_{\text{dim}}(\sigma) = \sum_{t=0}^{k-1} \text{MaxDim}(S_t).
\end{equation}
By prioritizing sequences with smaller maximum dimensions, this criterion focuses on reducing the most restrictive spatial constraint, enabling the robot to handle cartons more effectively within limited workspaces.

\begin{figure}
	\centering
	\includegraphics[height=3.2cm]{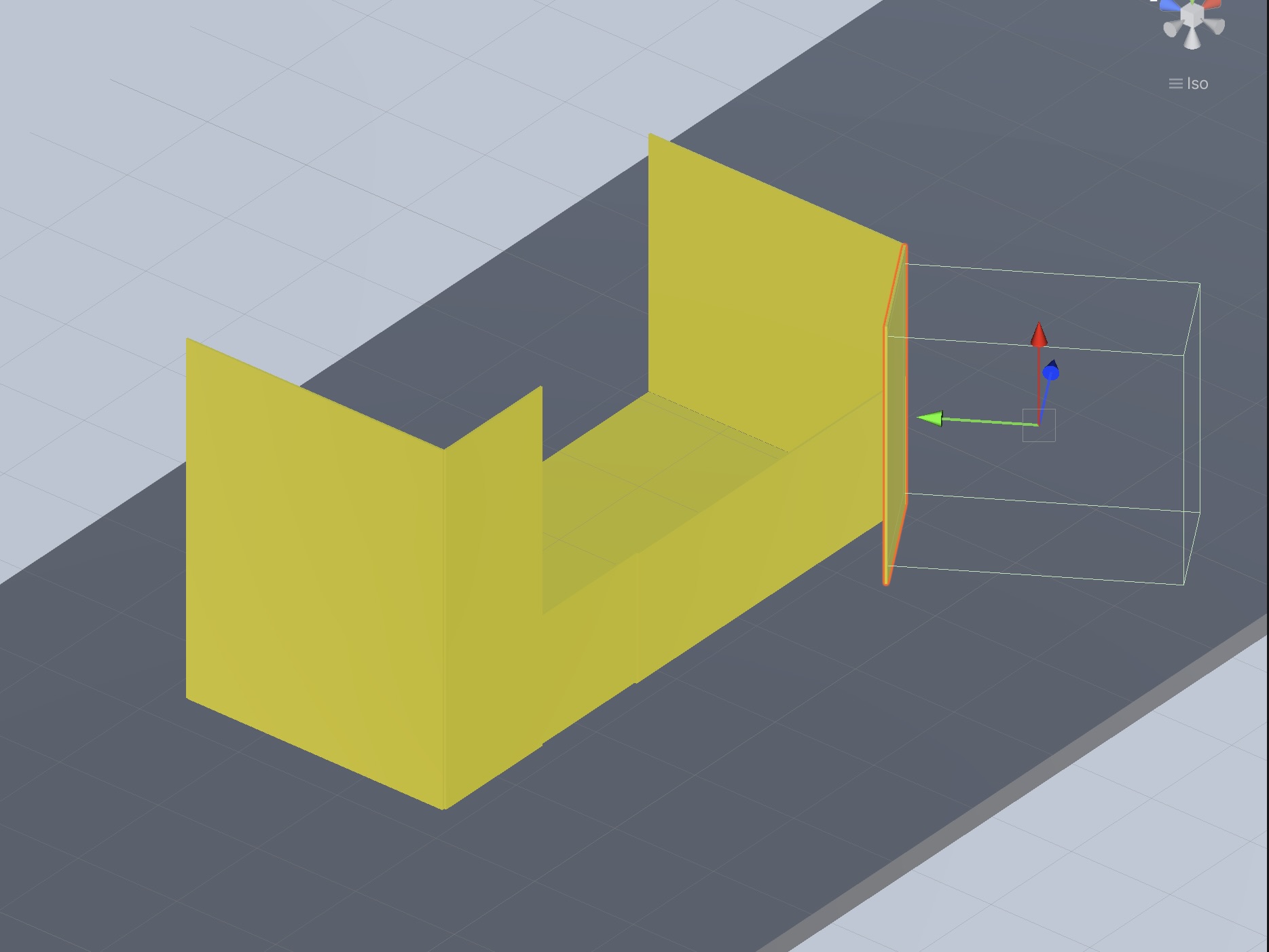}
	\includegraphics[height=3.2cm]{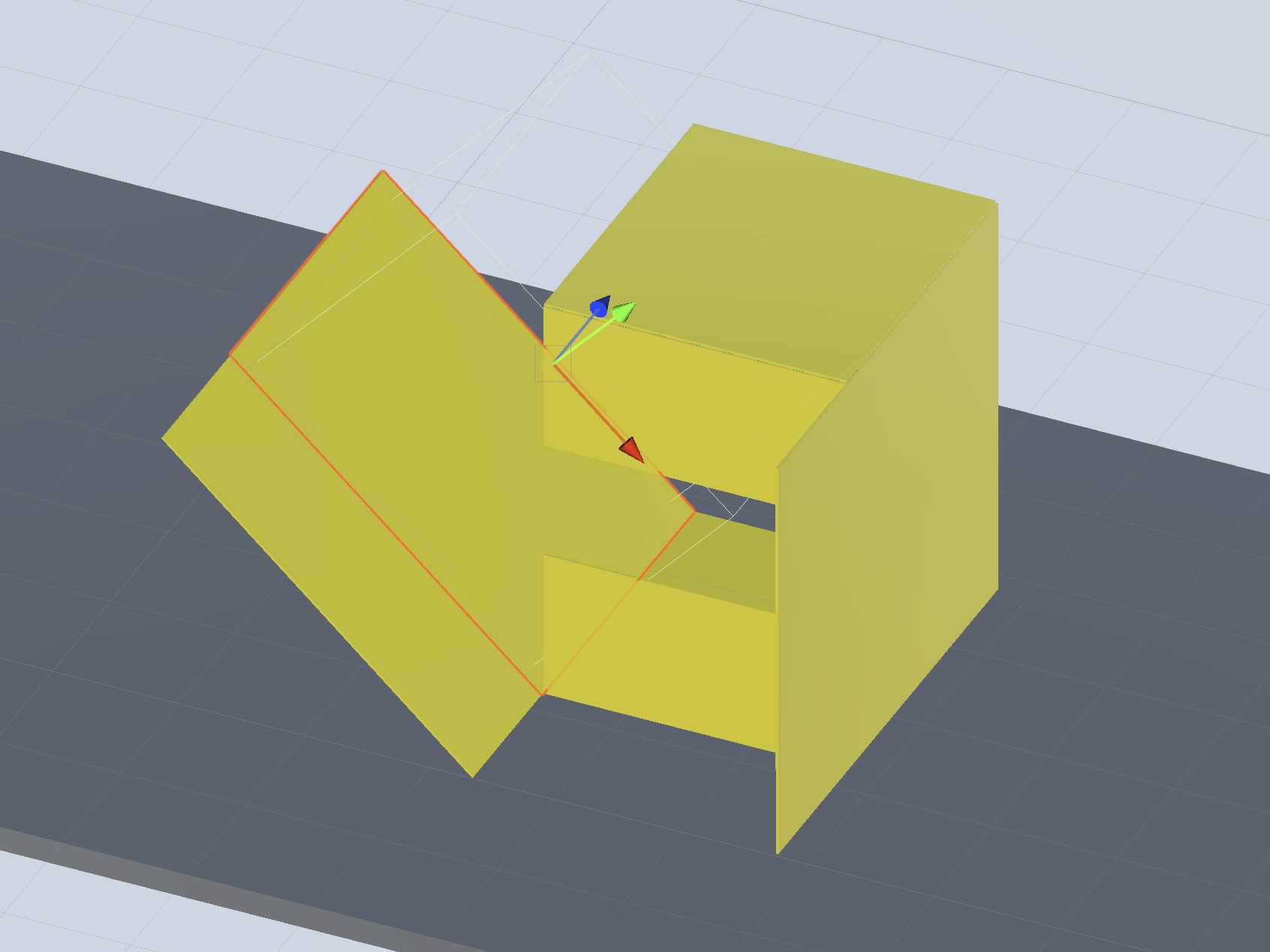}
	\caption{On the left, an example of \emph{Aerial Folding} is shown. On the right, the folding process involves a panel taken from the workbench, which does not classify as \emph{Aerial Folding}}
    \label{fig:aerial}
\end{figure}

\subsubsection{Number of Aerial Foldings Criterion}

This criterion evaluates the stability of the sequence by counting the number of folds performed without the support of the workbench denoting them as \emph{Aerial Folds}. An example is shown in Fig. \ref{fig:aerial}. Due to gravitational effects and material flexibility, these folds introduce variability in the initial position of the panel,  resulting in less predictable and repeatable folds.
For a given configuration $S_t$, an indicator function $a(S_t)$ is defined as:
\begin{equation}
a(S_t) = \begin{cases}
1 & \text{if an aerial fold occurs }, \\
0 & \text{otherwise}
\end{cases}
\end{equation}
The total cost $C_{\text{aerial}}(\sigma)$ for a sequence $\sigma$ is the cumulative sum of the indicator function:
\begin{equation}
C_{\text{aerial}}(\sigma) = \sum_{t=0}^{k-1} a(S_t).
\end{equation}
Minimizing $C_{\text{aerial}}(\sigma)$ enhances the stability and reliability of the folding process, especially in setups lacking external feedback sensors.
\begin{figure}
\begin{center}
\includegraphics[width=8cm]{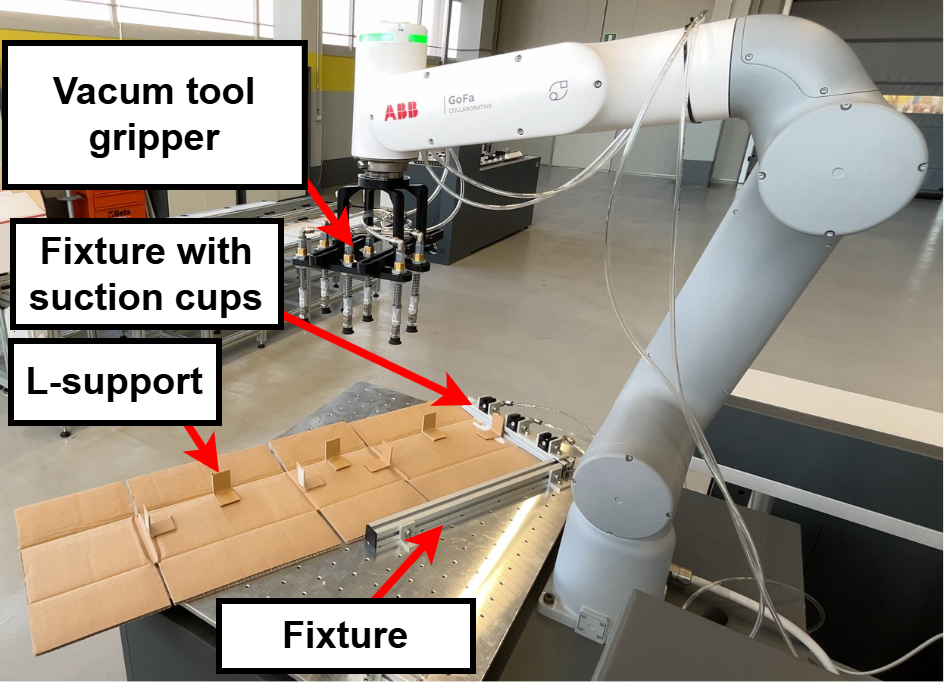}    
\vspace{-0.3cm}
\caption{Hardware setup used in our case study} 
\label{fig:hardware}
\end{center}
\end{figure}

\section{Implementation Overview and Case Study}
The following paragraph detail the implementation of the proposed method and provide a comprehensive example that demonstrates its application in a real-world scenario. First, we describe the technical implementation of the method, highlighting the steps taken to ensure accurate modeling, collision evaluation, and sequence extraction. Subsequently, a complete case study is presented, encompassing the hardware setup, the selection of a specific carton, and the execution of the folding sequence using the proposed approach.

\subsection{Simulation Carton Modeling}
The nature of the proposed approach heavily relies on precise collision control, a critical requirement for ensuring the feasibility of folding sequences. For this reason, the implementation was carried out with Unity, a simulation software capable of handling complex 3D simulations with detailed control over object interactions. The first phase of the implementation involves the representation of the carton, following the framework introduced in Section 3.1. The data required to fully describe the carton structure includes the precise positioning of panel centers, the dimensions of each panel, and the connectivity relationships that bind the panels together. This representation leverages the connectivity matrix to account for hereditary dependencies among panels.
Additionally, specific elements that influence the evaluation of collisions are incorporated into the simulation. These elements include the gripper, the table, and any supporting fixtures, as they play a significant role in determining the feasibility of folding actions. The algorithm uses this information not only to describe the spatial constraints but also to assess the accessibility of panels during the folding process. For instance, by incorporating the gripper dimensions, the algorithm can determine whether to grasp a panel from the inside or the outside.
\begin{figure}
\begin{center}
\includegraphics[width=8.3cm]{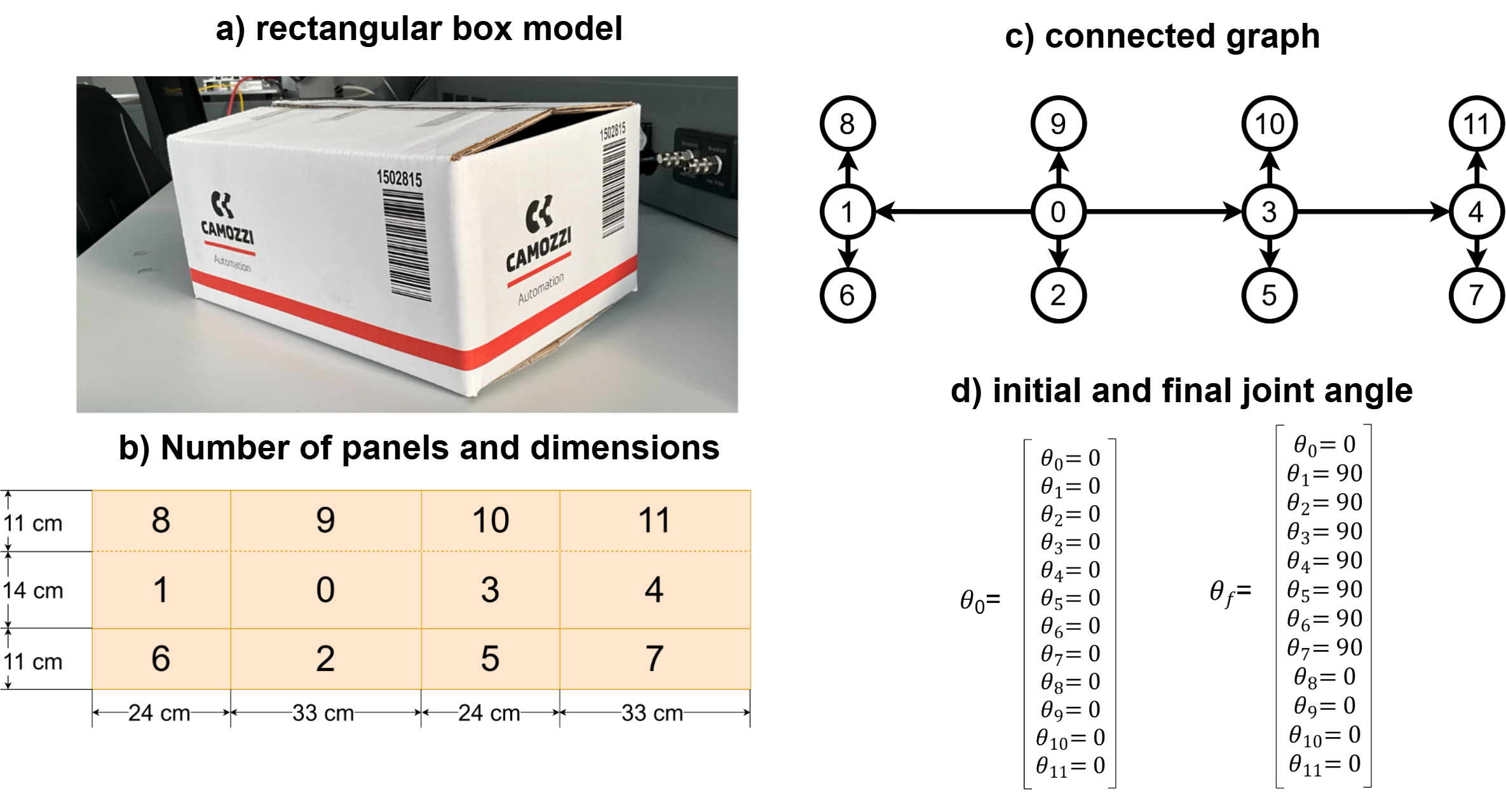}    
\vspace{-0.5cm}
\caption{Representation of the box model used in the use case, including all necessary parameters} 
\label{fig:example_box}
\end{center}
\end{figure}
Key parameters are defined to fine-tune the collision evaluation process. Among these, the tolerance angle specifies the granularity of collision checks, determining how frequently the algorithm evaluates potential collisions as the folding process transitions from the initial to the final state of each joint angle. Lastly, the initial and final angles of the carton panels are provided to describe their starting and target configurations.

\subsection{Case Study}
For the use case selected, a simple rectangular box model was chosen. The selected box has approximate dimensions of 33 cm in length, 24 cm in width, and 14 cm in height, with a panel thickness of 0.2 cm. As shown in Fig. \ref{fig:example_box}, the box is depicted with all the necessary parameters required to initiate the simulation.
The folding objective for this model focuses on assembling the base and lateral structure of the box while leaving the top open. This design choice allows for the potential insertion of items into the box before completing the final closing sequence.

\subsubsection{Experimental Setup}
The chosen setup significantly influences the metrics used to select the folding sequence for implementation. Specifically, the objective was to design a system capable of folding boxes using a relatively simple configuration: a single robotic manipulator equipped with modular fixtures and tools.
As can be seen from Fig. \ref{fig:hardware}, the tool is equipped with two rows of suction cups, which are essential for handling the carton panels. Each suction cup is mounted on a spring mechanism, providing flexibility to ensure a secure grip and maintain contact between the tool and the panel. The fixtures on the workbench include two orthogonal aluminum profiles mounted on its surface to guide the initial positioning of the flat carton. An additional row of suction cups is mounted above one of the aluminum profiles. These stabilizing suction cups are activated as needed to hold panels in place after a fold has been completed, preventing undesired movement and maintaining the integrity of the structure during the folding process.

Since the robot is designed to fold one panel at a time, it was necessary to implement a system to hold each panel in position after folding. To achieve this, the assembly procedure incorporates \emph{L-shaped supports} made from recycled carton material and covered with double-sided tape. These supports help secure the folded panels and reinforce the box structure. The robot places these supports automatically at designated spots along the folds by using one of the suction cups on the tool, which retrieves each support from an ideal dispenser positioned on the workbench.

The robot utilized in this setup is an ABB GoFa CRB 15000. A set of primitives designed for folding panels are parameterized with inputs derived from the simulation to execute the folding process effectively. This integration ensures that the robot movements align precisely with the recommendations generated during the simulation phase.

\subsubsection{Metrics for Folding Sequence Ranking}

The primary criterion selected was the minimization of \emph{aerial folds}. This criterion was chosen to prioritize stability and repeatability in practical implementation, especially given that the algorithm operates without vision feedback. In cases where multiple sequences resulted in the same number of aerial folds, a secondary criterion was applied: minimization of the \emph{maximum box dimension} during folding. This additional constraint, using a single robot, helps ensure that no single dimension of the partially folded carton exceeds the maximum distance reachable by the robot, which is limited in comparison to the fully extended size of the flat carton.
\begin{table}[H]
\begin{center}
\caption{First six sequences extracted by the algorithm with their associated weights}\label{table:seqExtr}
\begin{tabular}{lccc}
\hline
\textbf{Folding sequence} & \textbf{Volume} & \textbf{MaxDim} & \textbf{NAF} \\ \hline
{[2, 7, 4, 1, 3, 5, 6]} & 350052.2 & 604.0001 & 2 \\
{[2, 7, 4, 1, 3, 6, 5]} & 350052.2 & 604.0001 & 2 \\
{[2, 7, 1, 4, 3, 5, 6]} & 370809.8 & 594.5001 & 2 \\
{[2, 7, 1, 4, 3, 6, 5]} & 370809.8 & 594.5001 & 2 \\
{[2, 1, 7, 4, 3, 5, 6]} & 422120.9 & 561.2001 & 2 \\
{[2, 1, 7, 4, 3, 6, 5]} & 422120.9 & 561.2001 & 2 \\ \hline
\end{tabular}
\end{center}
\end{table}

\subsubsection{Sequence Results}
With these criteria in place, the simulation generated a ranked list of more than 100 valid folding sequences. This demonstrates the need for clear guidelines or metrics to suggest which of these sequences are worth implementing, ensuring their practical feasibility and alignment with the robot capabilities. The effective implementation of a sequence on the hardware is verified if the robot can utilize the sequence of the parametrized folding primitives without entering into collisions or exceeding the workspace. Table \ref{table:seqExtr} shows the first 6 ranked sequences with the related metrics. Having the same number of \emph{aerial folds} they have been ordered with respect to the \emph{maximum dimension} metrics.
By attempting to implement the sequences one by one, the first five were found to be not feasible due to fold starting points that were beyond the robot workspace. The sixth sequence, however, was feasible and was successfully implemented. The entire implemented sequence is shown in Fig. \ref{fig:folding}, starting with the flat box and ending with the final structure ready for object insertion, demonstrating the feasibility and reliability of the folding path.

\section{Conclusions and Future Works}
The primary objective of this work was the development of an algorithm capable of identifying and ranking feasible folding sequences for carton boxes. This algorithm was designed to address a specific gap in the literature by offering a flexible and efficient solution that generates all possible sequences, evaluates their validity through collision detection, and ranks them according to chosen criteria to meet hardware constraints. 
Besides the use case presented, more extensive validation will be conducted by extracting quantitative metrics to comprehensively validate the proposed approach. Specifically, statistical validation will be performed on multiple box types and through additional experiments to demonstrate that the current method consistently identifies a feasible solution in fewer attempts.
\begin{ack}
The authors acknowledge \emph{Camozzi Research Center} for providing robotic equipment, as well as for their support in this work.
\end{ack}

\bibliography{ifacconf}             
                                                   







\end{document}